\documentclass[sigconf]{acmart}
\usepackage{amsmath}
\usepackage{esvect}

\makeatletter
\@ifclasswith{acmart}{techreport}{
  \settopmatter{printacmref=false}
  \renewcommand\footnotetextcopyrightpermission[1]{}
  \fancyhead{}
  \fancyfoot{}
  \fancyhead[L]{\shorttitle}
  \fancyhead[R]{\shortauthors}
  \fancyfoot[C]{\thepage}
}{}
\makeatother

\begin{document}
\title{Human-in-the-Loop Refinement of Word Embeddings}

\author{James Powell , Kari Sentz , Martin Klein }
\affiliation{%
  \institution{Los Alamos National Laboratory}
  \city{Los Alamos}
  \state{New Mexico}
  \postcode{87545}
  \country{USA}
}
\email{{jepowell,ksentz,mklein}@lanl.gov}

\renewcommand{\shortauthors}{J. Powell et al.}
% This is a short software engineering paper about our adaptation and expansion of the retrofitting post-processing technique as an interactive tool for adjusting word embedding vectors in an embedding model. 

\begin{abstract}

Word embeddings are a fixed, distributional representation of the context of words in a corpus learned from word co-occurrences. Despite their proven utility in machine learning tasks, word embedding models may capture uneven semantic and syntactic representations, and can inadvertently reﬂect various kinds of bias present within corpora upon which they were trained. It has been demonstrated that post-processing of word embeddings to apply information found in lexical dictionaries can improve the semantic associations, thus improving their quality.  Building on this idea, we propose a system that incorporates an adaptation of word embedding post-processing, which we call "interactive refitting", to address some of the most daunting qualitative problems found in word embeddings. Our approach allows  a human to identify and address potential quality issues with word embeddings interactively. This has the advantage of negating the question of who decides what constitutes bias or what other quality issues may affect downstream tasks. It allows each organization or entity to address concerns they may have at a fine grained level and to do so in an iterative and interactive fashion. It also allows for better insight into what effect word embeddings, and refinements to word embeddings, have on machine learning pipelines.
\end{abstract}

\maketitle

\section{Introduction}

Word embeddings are a contemporary manifestation of the maxim
first posited by John Rupert Firth in 1957: “You shall know a word
by the company it keeps.” \cite{ref:Firth1957} When provided with an unlabeled text corpus as input, a single layer neural network outputs fixed length vectors of floating point values that represents each word's distinct context. Position and distance of word representations within this word embedding space encodes semantic and syntactic information.

There are two primary ways to use word embeddings -- intrinsically and extrinsically \cite{ref:WhitakerEtAl2019}. Intrinsic tasks use word embedding models to solve problems directly. For example, pair wise comparison of word embedding vectors using a distance metric can be used to find words that have similar meanings. Extrinsic tasks use word embeddings as a source of features that become input to machine learning pipelines, thus leveraging the information they encode in downstream tasks. This is an example of transfer learning, and has lead to rapid advancements in natural language processing.   
Selecting an appropriate word embedding model depends in large part on the intended use. In some instances, word embeddings trained on an extremely large corpus (also called global embeddings) are preferable due to the large vocabulary size and amount of training data used.  But in other cases, especially those involving specialized fields with distinct vocabularies, it may be necessary to find or train a local embedding model using a corpus of documents that use this specific vocabulary.

\section{Problem Statement}
Regardless of the source, most embedding models have quality
issues. Some embedding vectors represent multiple word senses, 
resulting in a diﬀuse representation that turns out not to represent any of the word senses very well. Others provide poor contextual 
representations simply because of a paucity of data, which is particularly a problem with local word embeddings. Sometimes an  embedding model may encode cultural biases, or be heavily influenced by false or misleading information sources, or it may skew toward particular opinions, political positions, or scientific disciplines.

Bias of all types--gender, racial, socioeconomic--is a well-documented problem in word embeddings, for example in  \cite{{ref:bolukbasietal2016}, {ref:BrunetEtAl2019}}. It is quite easy to find many examples with relatively trivial analogy queries against large embedding model such as Google News. It is therefore to be expected that some of the same biases would be found in local embedding models created with custom corpora. 

Word embeddings may also have problems evenly representing distinct scientific disciplines, particularly concepts that span multiple disciplines. A scientific corpus skewed towards physics and other hard sciences will generate better representations for terminology in those fields, but will offer little to no meaningful semantic representations for other fields such as sociology or linguistics. Nor would it be useful for identifying important co-occurrences that might be indicative of multidisciplinary eﬀorts. This problem, perhaps obviously, is more likely to occur for word embeddings generated for a scientifically focused corpus, and it exemplifies the tradeoﬀ between capturing a good representation for a technical vocabulary over modeling science more broadly. 

Digital libraries provide wide ranging access to intellectual and cultural artefacts. They are often at the cutting edge of theoretical and applied machine learning, particularly with regard to natural language processing and text. Machine learning models are used in digital libraries in many ways. They can be used to suggest topics for documents in a corpus, recommend or summarize content, perform automated text classification, and support query formulation and offer search term suggestions, to name a few. If a one of these services incorporates a word embedding model that contains encoded biases or skewed points of view, it could potentially produce unpredictable and potentially misleading or offensive output. This is broadly undesirable, and would be particularly harmful if for example, some type of bias affects the perception or distribution of historical or culturally sensitive content, or if it results in the inadvertent censoring of certain ideas due to current political sensibilities. It is also important to note that digital libraries can be the source of corpora used to produce word embedding models. As a team charged with developing services for digital libraries, both of these considerations motivated our interest in this problem.

\section{Prior Work}
Many techniques have been proposed for exploring word embeddings and the majority of them involve some type of visualization. Most utilize some type of dimensionality reduction rendered as a scatterplot, while some represent word vectors as graphs  \cite{{ref:smilkov2016}, {ref:katricheva2019}}. 

Various researchers have examined quality issues and possible solutions for word embeddings. Bolukbasi et al. \cite{ref:bolukbasietal2016} specifically examined the problem of gender bias in word embeddings, illustrating how this bias could be detected with analogy queries. They were able to identify geometric consistencies associated with gender bias and showed how this could form the basis of a debiasing algorithm that could reduce the impact of gender on otherwise gender-neutral words, while preserving the utility of the embedding model. Specifically, they noted that "there exists a low dimensional subspace in the embedding that empirically captures much of the gender bias." For gender bias, they were able to identify and remove vector component(s) that encode bias. Peng et al. \cite{ref:pengetal2021} examined various relationships among scientific disciplines captured by word embeddings trained on a variety of specific discipline literature, demonstrating that the embeddings encode commonly accepted relationships between scientific disciplines, as well as more nuanced relationships, but were problematic for characterizing multidisciplinary activities.

For other quality issues, a number of solutions have been proposed which combine detailed analysis of the embedding model with post-processing refinements, including  \cite{{ref:MuEtAl2017},{ref:MrksicEtAl2016},{ref:bolukbasietal2016}}.  Faruqui et al. \cite{ref:FaruquiEtAl2015} and  Mrk\v{s}i\'{c} et al. \cite{ref:MrksicEtAl2016} demonstrated that word embeddings can be improved by incorporating additional human-curated semantic information. The retrofitting technique identifies WordNet entries for terms in an embedding vocabulary and adjusts each so that it is closer to its synonyms.

Given the circumstances surrounding word embeddings and
the findings of these and other researchers regarding the potential
for selectively mitigating embedding quality issues, we chose to
investigate the viability of an interactive solution to this problem. We believe that a successful approach needs to incorporate two major capabilities: interactive exploration of an embedding space, and the ability to make live adjustments to embedding vectors it contains. The remainder of this paper examines this idea.

\section{Dataset}

For all of the experiments described below, we used the publicly available word2vec model for Google News, which was trained on 100 billion words. This allowed us to validate capabilities of our prototype by exploring and replicating selected issues identified in prior work, such as occurrences of gender bias.

\section{System Description}
Our prototype system is designed to allow for interactive, iterative exploration and modification of any compatible word embedding model. It emphasizes a four phased approach: "search - navigate - visualize - adjust". All four phases are in service of the goal of surfacing quality issues in word embeddings and making it possible to address these issues directly and immediately within the model. We implemented a prototype of this concept, at the core of which is a capability we refer to as interactive refitting. The prototype consists of the following components:

\begin{itemize}
\item{A search interface for providing one or more search terms }
\item{A textual result interface that presents most similar terms from the word embedding model}
\item{Links to explore these results via three Web visualization tools}
\item{A mechanism allowing users to identify and specify target for adjustment via refitting }
\item{A Web services wrapper for the retrofitting objective}
\item{An update function that modifies selected word2vec embedding vectors and stores them in the live model}
\end{itemize}

\section{Experiments}
\begin{figure}
    \centering
    \includegraphics[width=.65\linewidth]{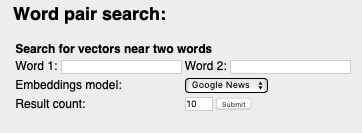}
    \caption{Search embedding space using mean of two word vectors}
    \label{fig:embedsearchwordpair}
\end{figure}

We first test our prototype's discovery capabilities and found that we could readily locate examples of direct bias with basic word searches. In word embeddings, direct bias can be identified when a gender-neutral word is measurably closer to pronouns representing one gender as opposed to another. In our prototype, this is accomplished by searching for examples of male and female pronouns, and a gender neutral word, and comparing the distance values. This is accomplished with a two term query. The embedding vectors for each term is used to find common neighbors (Figure ~\ref{fig:embedsearchwordpair}). For example, a query for "he" and "nurse" returned "she" as the nearest term to these two vectors, while "she" and "nurse" returned "registered nurse" as the nearest term. Table ~\ref{tab:analogy} shows the top five nearest matches for these two queries.  

\begin{center}
\small
    \begin{tabular}{ |c|c|c|c|}
    \hline

\hline
\multicolumn{2}{|c|}{\textbf{"he" + "nurse"}} & \multicolumn{2}{|c|}{\textbf{"she" + "nurse"}} \\
\hline
doctor & .6049 & registered nurse & .6949 \\
\hline
registered nurse & .5889 & neonatal nurse & .6094 \\
\hline
x ray technician & .5814 & woman & .6076 \\
\hline
medic & .5507  & nurse practitioner & .6057 \\
\hline
candy striper & .533 & mother & .6048 \\
\hline

\end{tabular}
\captionof{table}{Analogy occupation query}
\label{tab:analogy}
\end{center}

So, for the pronoun "he," in combination with the occupation
"nurse", the nearest term is "doctor", while for "she" + "nurse", it is "registered nurse." As the remainder of the results show, there is a strong tendency for nursing occupations to be associated with  female gender pronouns, while "doctor," "x-ray technician." and "medic" fill three of the top five slots for "he" + "nurse" in the same embedding space. Other examples include for "she" versus "he" and "bookkeeper": "homemaker" appears in the top five for the female gender combination, while "accountant" appears in the top five results for male. However, "she" + "hairdresser" results in "beautician" while "he" + "hairdresser" is closest to "barber."  While "beautician" and "barber" might be viewed by some as outdated and sexist, it is notable that the term "hair stylist" is also in the top three closest terms for both combinations. In this case, the actual occupation is identical, the diﬀerence is merely between  gendered labels for the same job. This demonstrates the value of being able to apply human judgment to evaluating potential bias in a corpus. 

A second way to explore direct gender bias is to look for the presence of gender stereotypes via analogy, which is another form of search supported by our prototype.  An example of an extreme stereotype identified in \cite{ref:bolukbasietal2016} was "he-she"/ "snappy-sassy." When we performed an analogy query  \[\vv{\text{sassy}}-\vv{\text{she}}+\vv{\text{he}}\] the top five closest matches were  "swaggering", "cocky", "suave", "brash" and "genial."  The top five words for a related query  \[\vv{\text{snappy}}-\vv{\text{he}}+\vv{\text{she}}\] 
were "sassy" "perky" "cutesy"  "bitchy" and "saucy." In both cases,
some of the matches were somewhat negative, but in our view,
those returned for the female analogy query tended to be more negative. This is admittedly a subjective judgement, but the goal of our interactive prototype is to allow individuals to discover and asses potentially problematic word relationships. 

\begin{figure}
    \centering
    \includegraphics[width=.65\linewidth]{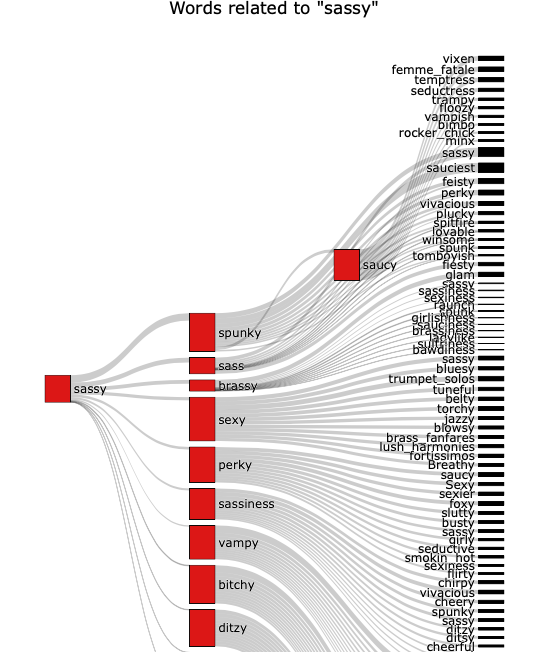}
    \caption{Embedding search results as sankey diagram}
    \label{fig:embedsearchsankey}
\end{figure}

As these results show, we were able to demonstrate that a user could perform simple text searches and find examples of bias as well as other potential quality concerns in a word embedding model. Our prototype currently supports four query types: single term query, additive search for two terms, subtractive search, and analogy search. Results are presented by default as an n-nearest terms list ordered by cosine distance. To enhance the comprehensibility of the results, we added several visualization options, including sankey diagrams (illustrated in Figure ~\ref{fig:embedsearchsankey}), force directed graphs, t-SNE plots, and heatmaps.

\section{Interactively refining word embeddings}
Discovering problems in a word embedding model is just the first step, the next is providing a technique for making improvements. We propose a human-in-the-loop approach for refining word embeddings, which we call interactive refitting. Our implementation adapts the approach described in \cite{ref:FaruquiEtAl2015} to accommodate  interactive user input. We  implemented two forms of refitting: 1) an interactive version of the original retrofitting objective, where the user specifies a list of terms and a target term. This results in the target being moved closer to the group, and  2) a round-robin style refitting, where the user identifies a group of terms that should be moved closer to one another. 

\begin{figure}
  \includegraphics[width=.7\linewidth]{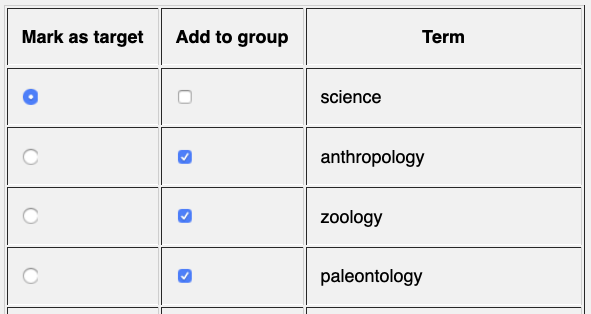}
  \captionof{figure}{Protoype refitting UI}
  \label{fig:retrofitui}
\end{figure}

Interactive refitting changes the embedding vectors representing one or more terms, and these changes are applied to the embedding model in real time. This provides two benefits: model updates are immediately available to downstream applications, and the changes are immediately apparent when the user performs new searches.

Here we demonstrate the effects of the two forms of interactive refitting with the goal of improving the representation of "science" in the Google News embedding model. Starting from an unmodified version of this model, we first search for the term "physics". The results are dominated by more specific physics-related terms including theoretical physics, quantum mechanics, and astrophysics. We would like to move "physics" a bit closer to other terms that represent other scientific disciplines. From the search results, we  select five terms we would like to apply refitting to: astronomy, biochemistry, biology, biophysics, and science. These five terms are combined with the original search term "physics" and a refitting request is initiated. This adjusts the embedding vector for each term so they are all moved closer to one another. Table ~\ref{tab:physics} shows how the five selected terms changed in relation to "physics."

\smallskip
\begin{center}
\small
    \begin{tabular}{|c|c|c|c|}
    \hline
{\textbf{Term}} & {\textbf{Before}} & {\textbf{After}} \\
\hline
\hline
science & .5929 & .8663 \\
\hline
astronomy & .5644 & .8379 \\
\hline
biophysics & .5623 & .8399 \\
\hline
biology & .5387 & .8428 \\ 
\hline
biochemistry & .5305 & .8389 \\
\hline
\end{tabular}
\captionof{table}{Moving the term "physics" closer to other disciplines}
\label{tab:physics}
\end{center}

\smallskip
\begin{center}
\small
    \begin{tabular}{|c|c|c|c|}
    \hline
{\textbf{Term}} & {\textbf{Before}} & {\textbf{After}} \\
\hline
\hline
anthropology & .6308 & .8661 \\
\hline
zoology & .6158 & .8306 \\
\hline
paleontology & .5909 & .7893 \\
\hline
psychobiology & .5777 & .7319 \\ 
\hline
developmental psychology & .5774 & .7307 \\
\hline
\end{tabular}
\captionof{table}{Cosine distance between "science" and selected terms, before and after refitting}
\label{tab:target-science}
\end{center}

For the second test, we perform targeted refitting . This is an interactive equivalent to post-processing retrofitting described in \cite{ref:FaruquiEtAl2015}, which moves a single target term toward its synonyms. In this case, we search for "science." As the first two columns of Table ~ref{tab:science} show, the top ten closest terms to "science" were a decidedly mixed bag. So we select "science" as our target and a group of five terms to which "science" should be moved toward, and initiate a second refitting action. Table ~\ref{tab:target-science} shows the results of this action.

\begin{figure}
  \includegraphics[width=1.1\linewidth]{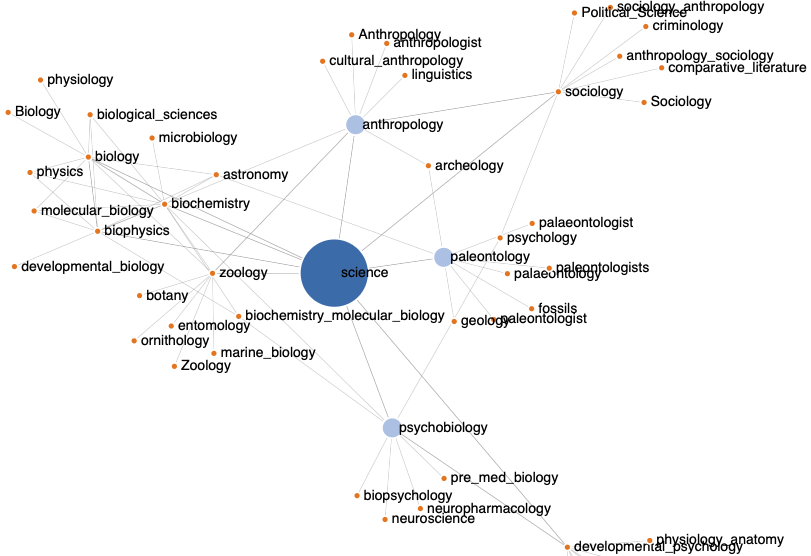}
  \captionof{figure}{Graph visualization illustrating the improvements related to the term "science" after refitting}
  \label{fig:science-after}
\end{figure}

After two refitting actions, we perform a search for "science" for a second time. The changes to the representation of "science" in the embedding model are dramatic, and we think far more reasonable. The changes are illustrated in Table ~\ref{tab:science} and in Figure ~\ref{fig:science-after}.

\medskip
\begin{center}
\small
    \begin{tabular}{|c|c|c|c|}
    \hline
\multicolumn{2}{|p{2cm}|}{\textbf{Before refitting}} &  \multicolumn{2}{|p{2cm}|}{\textbf{After refitting}} \\
\hline
\hline
faith Jezeierski & .6965 &  anthropology & .8661 \\ 
\hline
sciences &.6821 &  zoology & .8306\\
\hline
biology & .6775 &  paleontology & .7892 \\
\hline
scientific & .6535 &  biology & .7868 \\
\hline
mathematics & .63 &  biochemistry & .7778 \\ 
\hline
Hilal Khasan professor & .6153 &  psychobiology & .7319 \\
\hline
impeach USADA & .6149 & developmental psychology & .7305 \\
\hline
professor Kent Redfield & .6144 & biophysics & .7284  \\
\hline
physics astronomy & .6105 & sociology & .7153 \\
\hline
bionic prosthetic fingers & .6083 &  biological anthropology & .7149 \\ 

\hline
\end{tabular}
\captionof{table}{Terms closest to "science" in the Google News embedding model, before and after refitting}
\label{tab:science}
\end{center}

\section{Conclusion and Future Work}
There is a growing awareness of the ease with which machine learning algorithms and models can learn to model negative stereotypes, favor false ideas, and incorporate biases that have historically caused great harm, even though there is no societal value in allowing them to do so. Word embeddings often manifest these problems. We have described a proof-of-concept system that allows humans to use their own intuition and judgement to identify and neutralize these kinds of problems. We believe this is a viable approach for several reasons: 1) qualitative issues with word embeddings are often highly specific to given machine learning task, 2) issues can assume many forms including bias, skewing, or poor quality due to data sparsity, and 3) some of these issues can only be effectively identified by a human. 

In the future, we would like to assess and characterize the range of quality issues that could be discovered and mitigated using this technique. We would also like to conduct a quantitative assessment and comparison of this approach as compared to non-interactive data-intensive analytical approaches. We plan to investigate adding additional refitting capabilities such as allowing users to move words away from one another in addition to toward each other. We would like to determine if our interactive solution can be used to introduce broad qualitative and consistent improvements to word embeddings.  We would also like to investigate how well this technique scales, and in what context it might it be introduced. Finally we would like to test embedding models that have been modified using this technique to see what the effect is on extrinsic machine learning tasks. 

\bibliography{acl2018}
\bibliographystyle{acl_natbib}

\end {document}